\documentclass[preprint,12pt]{elsarticle}

\pdfoutput=1

\usepackage{amssymb}

\usepackage{amsmath,epsfig}
\usepackage{svg} 
\usepackage{adjustbox}
\usepackage{setspace}
\usepackage{makecell}
\usepackage{multirow}
\usepackage{graphicx}
\usepackage{longtable}  
\usepackage{float}    

\begin{document}

\begin{frontmatter}



\title{A Timely Survey on Vision Transformer for Deepfake Detection}


\author{Zhikan Wang$^{1,2}$, Zhongyao Cheng$^{1}$, Jiajie Xiong$^{1,2}$,  Xun Xu$^{1}$, Tianrui Li$^{3}$, Bharadwaj Veeravalli$^{2}$, Xulei Yang$^{1}$}


\address{
$^1$Institute for Infocomm Research (I$^2$R), A*STAR,  Singapore\\
$^2$National University of Singapore (NUS), Singapore\\
$^3$Southwest Jiaotong University (SWJTU),  China}

\begin{abstract}
In recent years, the rapid advancement of deepfake technology has revolutionized content creation, lowering forgery costs while elevating quality. However, this progress brings forth pressing concerns such as infringements on individual rights, national security threats, and risks to public safety. To counter these challenges, various detection methodologies have emerged, with Vision Transformer (ViT)-based approaches showcasing superior performance in generality and efficiency. This survey presents a timely overview of ViT-based deepfake detection models, categorized into standalone, sequential, and parallel architectures. Furthermore, it succinctly delineates the structure and characteristics of each model. By analyzing existing research and addressing future directions, this survey aims to equip researchers with a nuanced understanding of ViT's pivotal role in deepfake detection, serving as a valuable reference for both academic and practical pursuits in this domain.
\end{abstract}

\begin{graphicalabstract}
\includegraphics[width=1\textwidth]{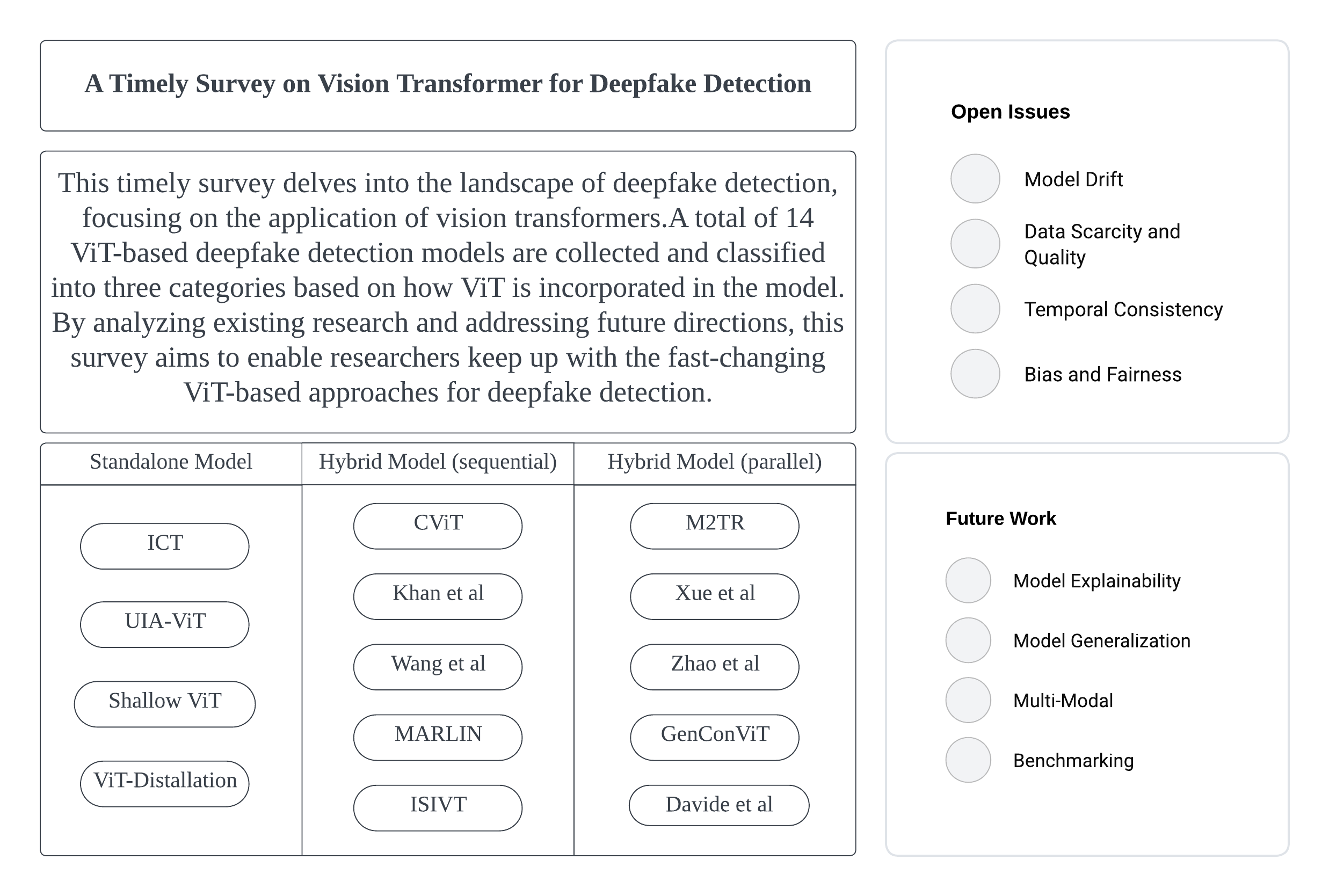}

\end{graphicalabstract}

\begin{highlights}
\item This survey identifies ViT-based deepfake detection as a cutting-edge research direction, highlighting the need for a timely literature review to keep up with rapidly evolving ViT-based approaches.
\item This survey provides an up-to-date overview (as of February 28, 2024) of a total of 14 ViT-based deepfake detection models categorized into standalone, sequential, and parallel architectures.
\item This survey also explores unresolved issues in deepfake detection and suggests potential directions for future research.
\end{highlights}


\end{frontmatter}


\section{Introduction}
Deepfakes involve the generation of images or videos that convincingly depict events or individuals who may not exist or have never engaged in the portrayed activities. These sophisticated manipulations of multimedia content, often driven by deepfake generative models, have the potential to mislead, manipulate, and pose significant threats to a variety of areas, including politics~\cite{politic}, media~\cite{media}, and personal privacy~\cite{social}. Driven by advances in deep learning, deepfake has emerged as a more powerful tool for creating highly realistic synthetic media, raising significant concerns about the potential misuse and manipulation of visual content, which underscores the importance of the development of efficient detection mechanisms.

To address the problems posed by deepfake, various detection methods have been proposed. Traditional techniques for detecting deepfakes utilize image and video analysis methods, such as local feature extraction and motion analysis, to identify manipulated content. More recently, a variety of deep learning methodologies, including long short-term memory (LSTM), recurrent neural network (RNN), and hybrid approaches, have been introduced for the detection of manipulated images and videos~\cite{survey2}. Within the diverse array of existing methodologies, approaches built on Vision Transformer (ViT) have emerged as state-of-the-art in current research.

This timely survey delves into the landscape of deepfake detection, focusing on the application of vision transformers~(ViTs)~\cite{ViT} – a novel neural network architecture that have demonstrated remarkable success in computer vision tasks. ViTs depart from the traditional convolutional neural networks (CNNs) by building upon self-attention mechanisms, allowing them to capture global dependencies in data sequences~\cite{globalvit}. The comprehensive modeling of global context by ViTs enhances its efficacy in capturing overarching features and relationships within images. This capability is proven advantageous for discerning subtle details and characteristics within an image. Consequently, in the domain of deepfake detection, the utilization of ViTs presents a distinctive avenue.

The motivations behind this survey are multifaceted. Firstly, deepfake detection is crucial for safeguarding the integrity of visual content and mitigating the potential harm caused by deceptive multimedia. Secondly, the unique characteristics of ViTs, with their attention-based architectures, offer a promising paradigm for addressing the challenges inherent in deepfake detection. Finally, ViT-based deepfake detection is a cutting-edge research direction, with many researchers working in parallel. There is a need for a timely literature review to keep up with the fast-changing ViT-based approaches to enable researchers to understand ViT's features better and develop robust strategies. This survey aims to provide a comprehensive overview of existing ViT-based deepfake detection models, their strengths, limitations, open issues, and future directions.

The rest of the article is organized as follows: In Section 2, we provide a basic introduction pertaining to the generation and detection of deepfakes, along with a navigation of related surveys. Section 3 is devoted to ViT-based approaches for detecting deepfakes. Furthermore, Section 4 delves into open issues and potential directions for future research. The survey concludes with a summary in Section 5.

\section{Related Works}

\subsection{Deepfake Generation}
Deepfakes are generated through the manipulation of pre-existing videos and images, resulting in the creation of content that exhibits a convincing semblance of authenticity despite being entirely synthetic. Deepfake generation can be categorized into four main groups: identity swap, face reenactment, attribute manipulation, and entire face synthesis~\cite{DeepfakeSurvey}.
Identity swap encompasses the substitution of an individual's facial features in an image or video with those of another person. Face reenactment entails the modification of the facial expressions exhibited by an individual within an image or video. In the domain of attribute manipulation deepfakes, there is the manipulation of specific facial attributes, including skin tone, age, gender, and the incorporation or removal of elements like glasses. The synthesis of entire faces involves the generation of facial samples portraying individuals that do not exist in reality~\cite{juefeixu2022countering}.

\subsection{Traditional Methods for Deepfake Detection}
Traditional approaches to tackling deepfake-related challenges often refer to methods and algorithms developed before the advent of advanced deep learning models. These methods typically involve manual inspection, forensic analysis~\cite{9078989}, or rule-based approaches~\cite{Agarwal2019ProtectingWL} to identify anomalies and inconsistencies within multimedia content. Various classic pattern recognition methods have been utilized for deepfake detection, encompassing logistic regression, probabilistic linear discriminant analysis, random forest, gradient boosting decision tree, extreme learning machine, k-nearest neighbor, support vector machine, Gaussian mixture model, etc~\cite{yi2023audio}. Although these approaches may provide some level of effectiveness, their limitations become more evident as deepfake technologies become increasingly complex and sophisticated. Consequently, contemporary efforts have shifted towards harnessing advanced deep learning techniques, such as deep neural networks and ViTs, to improve the accuracy and efficiency of deepfake detection. 
\subsection{Deep Learning for Deepfake Detection}
Deep learning, a machine learning technique based on neural networks, has gained significant attention in recent years due to its remarkable achievements in diverse fields. It has also shown promising results in deepfake detection. CNN~\cite{10.1145/3267357.3267367}, RNN~\cite{sabir2019recurrent}, LSTM~\cite{9497385}, and ViT are the most widely used deep learning techniques in deepfake detection. CNNs excel in capturing spatial dependencies within images, making them particularly effective in discerning facial manipulations and other visual anomalies in frame-level detection. In video-level detection, RNNs and LSTMs are well-suited for the sequential video frames. Their recurrent architectures enable the modeling of temporal dependencies and can capture subtle changes over time (e.g., facial expressions, eye blinks, lip movement). ViTs enhance the understanding of the overall structure of an image, facilitating the identification of inconsistencies or anomalies that may be indicative of manipulation.

\subsection{ViT-based Deepfake}
Transformers, primarily applied to high-level vision tasks, face challenges in low-level vision tasks like image super-resolution due to the complexity of detailed image generation~\cite{SurveyTrans}. ViTs primarily focus on learning global representations and might lack the fine-grained modeling required for generating realistic and highly detailed facial features necessary for deepfakes, their application in generating detailed features for tasks like deepfake generation is limited. While originally designed for vision-related tasks like image recognition, ViTs offer unique advantages in analyzing and understanding the intricate details of deepfake images and videos. By leveraging the power of transformers and attention mechanisms, ViTs provide a new avenue for enhancing deepfake detection algorithms and advancing the field of multimedia forensics. In this article, we explore the application of ViTs in deepfake detection, highlighting their potential, advantages, and challenges in combating the proliferation of manipulated media content. The details of ViT-based deepfake detection techniques will be discussed in Section 3.
\subsection{Surveys on Deepfake}
In the years since deepfake appeared, a number of surveys have been conducted on deepfake in the literature. The authors of~\cite{DeepfakeSurvey} reviewed the theoretical concepts, foundation, and classification of the deepfake for both generation and detection. A similar study, ref.~\cite{survey1}, discussed deep fake generation, detection, datasets, challenges, and research directions. In~\cite{survey2} and~\cite{nguyen2022deep}, an in-depth exploration of the application of deep learning in deepfake is given. Yi et al. in~\cite{yi2023audio} provided a more nuanced perspective of investigation, focusing on the audio deepfake detection.~\cite{SEOW2022351} provided a comprehensive overview of deepfake and underlined publicly available deepfake generation tools and datasets for benchmarking. In addition,~\cite{ieee1},~\cite{ieee2},~\cite{ieee3} have also investigated deepfake to varying degrees. In spite of the numerous surveys conducted on the subject of deepfake, there is a notable scarcity of surveys that address the deepfake technology associated with ViTs. Our survey serves as a valuable supplement in this specific direction, providing researchers with a more focused and nuanced reference in the realm of deepfake technology, particularly pertaining to ViTs.
\begin{figure}[tb]
    \centering
    \includegraphics[width=0.9\textwidth]{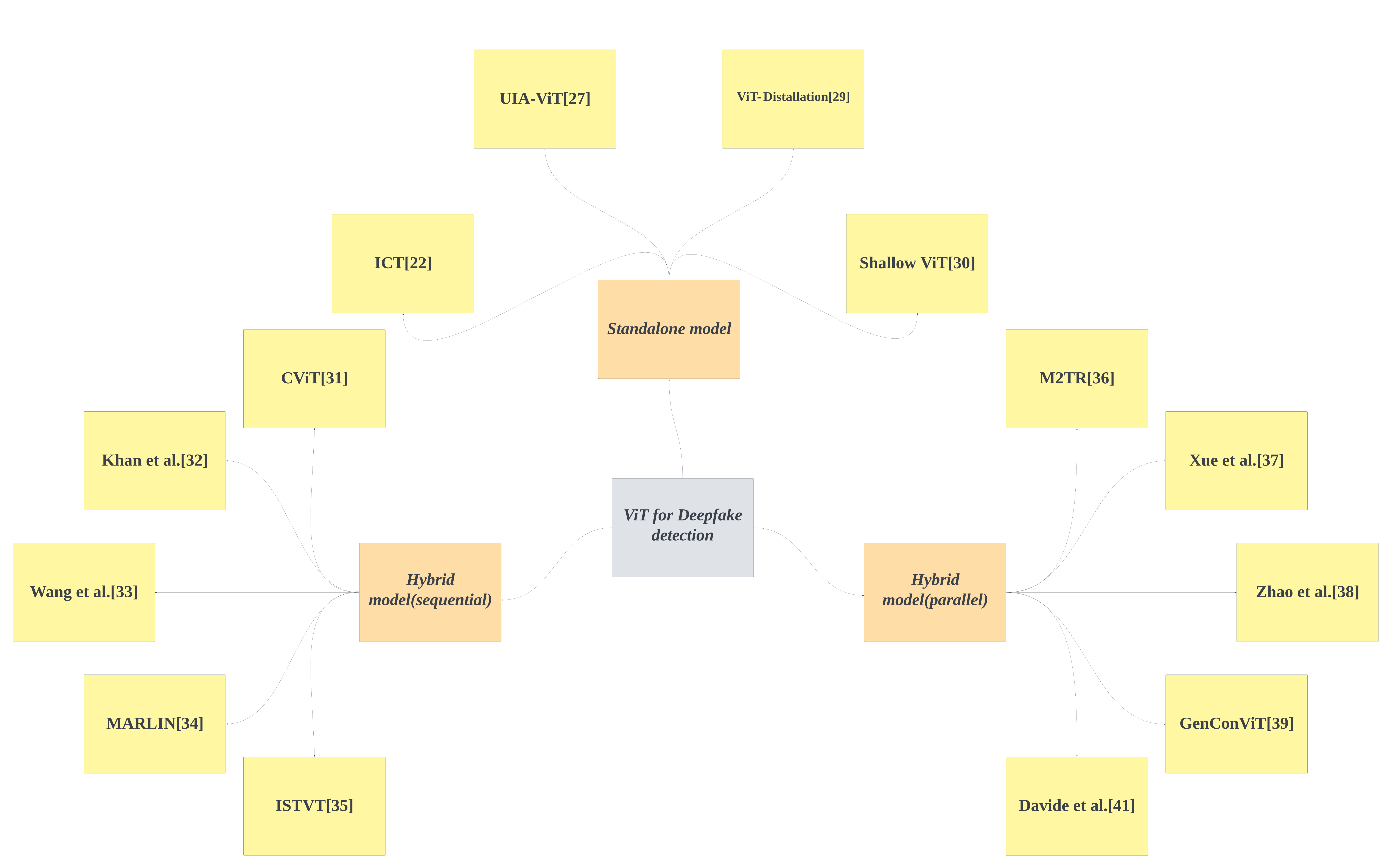}  
    \caption{An overview of the main models discussed in this survey.}	
    \label{overview}
\end{figure}

\begin{table*}
    \centering
    \begin{adjustbox}{max width=\textwidth}
    \begin{tabular}{c c c c c}
        \hline
        \textbf{Model} & \textbf{Dataset} & \textbf{Performance} & \textbf{Source Code (accessed on 8 January 2024)} & \textbf{Year} \\
        \hline
        \multicolumn{5}{c}{} \\
        \multicolumn{5}{c}{\textbf{Standalone model}} \\
        \multicolumn{5}{c}{} \\
        \hline
        \multirow{2}{*}{ICT~\cite{ICT}} & \makecell{FaceForensics++~\cite{Mesoscopic90},\\ Deeper~\cite{jiang2020deeperforensics10},\\  Celeb-DF-v1~\cite{li2020celebdf},\\  Celeb-DF-v2~\cite{li2020celebdf},\\DFD~\cite{DFD}} & \makecell{AUC= 90.22 \\ AUC= 93.57 \\ AUC= 81.43 \\ AUC= 85.71 \\AUC=84.13} & \makecell{https://github.com/LightDXY/ \\ } & \multirow{2}{*}{2022} \\
        \hline
        \multirow{2}{*}{UIA-ViT~\cite{UIA-ViT}} & \makecell{FaceForensics++(HQ)~\cite{Mesoscopic90},\\ DFD~\cite{DFD},\\ Celeb-DF-v2~\cite{li2020celebdf},\\ Celeb-DF-v1~\cite{li2020celebdf},\\ DFDC~\cite{dolhansky2020deepfake}} & \makecell{AUC=99.33 \\ AUC*=94.68 \\ AUC*=82.41 \\ AUC*=86.59 \\ AUC*=75.80} & \makecell{NA} & \multirow{2}{*}{2022} \\
        \hline
        \multirow{1}{*}{ViT-Distillation~\cite{Distillation}} & \makecell{DFDC~\cite{dolhansky2020deepfake}} & \makecell{AUC=97.80, f1=91.9} & \makecell{https://github.com/smu-ivpl/DeepfakeDetection} & 2021 \\
        \hline
        \multirow{2}{*}{Shallow ViT~\cite{Shallow}} & \makecell{RFF~\cite{Shallow},\\ DRFFD~\cite{Shallow}} & \makecell{ACC=92.15, AUC=92.00 \\ ACC=88.52, AUC=88.00 } & \makecell{NA} & \multirow{2}{*}{2022} \\
        \hline
        \multicolumn{5}{c}{} \\
        \multicolumn{5}{c}{\textbf{Hybrid model (Sequential)}} \\
        \multicolumn{5}{c}{} \\
        \hline
        \multirow{2}{*}{CViT~\cite{
CVIT}} & \makecell{FaceForensics++(HQ)~\cite{Mesoscopic90},\\ DFDC~\cite{dolhansky2020deepfake}} & \makecell{ACC=93.75 \\ ACC=91.5, AUC=91 } & \makecell{https://github.com/erprogs/CViT \\} & \multirow{2}{*}{2021} \\
        \hline
        \multirow{2}{*}{Khan et al.~\cite{Increment}} & \makecell{FaceForensics++(HQ)~\cite{Mesoscopic90},\\ DFD~\cite{DFD},\\ DFDC~\cite{dolhansky2020deepfake}} & \makecell{ACC=99.79 \\ ACC=99.28 \\ ACC=91.69} & \makecell{https://github.com/sohailahmedkhan/ \\ } & \multirow{2}{*}{2021} \\
        \hline
        \multirow{2}{*}{Wang et al.~\cite{Pooling}} & \makecell{FaceForensics++~\cite{Mesoscopic90},\\ DFDC~\cite{dolhansky2020deepfake},\\ Celeb-DF~\cite{li2020celebdf},\\DF-1.0~\cite{jiang2020deeperforensics10}} & \makecell{ACC=92.11, AUC=97.66 \\ ACC*=65.76, AUC*=73.68 \\ ACC*=63.27, AUC*=72.43 \\ ACC*=62.46, AUC*=78.19} & \makecell{NA} & \multirow{2}{*}{2023} \\
        \hline
        \multirow{1}{*}{MARLIN~\cite{cai2023marlin}} & \makecell{FaceForensics++~\cite{Mesoscopic90}} & \makecell{ACC=90.71,AUC=93.77} & \makecell{NA} & \multirow{1}{*}{2023} \\
        \hline
        \multirow{1}{*}{ISTVT~\cite{ISTVT}} & \makecell{FaceForensics++(HQ)~\cite{Mesoscopic90}\\FaceForensics++(LQ)~\cite{Mesoscopic90}\\Celeb-DF~\cite{li2020celebdf},\\DFDC~\cite{dolhansky2020deepfake}} & \makecell{ACC=99.0\\ACC=96.15\\ACC=99.8\\ACC=92.1} & \makecell{NA} & \multirow{1}{*}{2023} \\
        \hline
        \multicolumn{5}{c}{} \\
        \multicolumn{5}{c}{\textbf{Hybrid model (Parallel)}} \\
        \multicolumn{5}{c}{} \\
        \hline
        \multirow{2}{*}{M2TR~\cite{M2TR}} & \makecell{FaceForensics++(LQ)~\cite{Mesoscopic90}, \\ FaceForensics++(HQ)~\cite{Mesoscopic90}, \\ FaceForensics++(RAW)~\cite{Mesoscopic90}, \\ Celeb-DF~\cite{li2020celebdf}, \\ SR-DF~\cite{M2TR}} & \makecell{ACC=92.89, AUC=95.31 \\ ACC=97.93, AUC=99.51 \\ ACC=99.50, AUC=99.92 \\ AUC=95.5 \\ AUC=86.7} & \makecell{https://github.com/wangjk666/ \\} & \multirow{2}{*}{2022} \\
        \hline
        \multirow{2}{*}{Xue et al.~\cite{FOFDTD}} & \makecell{FaceForensics++(LQ)~\cite{Mesoscopic90}, \\ FaceForensics++(HQ)~\cite{Mesoscopic90}, \\ FaceForensics++(RAW)~\cite{Mesoscopic90}, \\ DFD~\cite{DFD}, \\ DFDC~\cite{dolhansky2020deepfake}, \\ Celeb-DF~\cite{li2020celebdf}} & \makecell{ACC=94.14, AUC=96.43 \\ ACC=98.12, AUC=99.67 \\ ACC=99.67, AUC=99.93 \\ AUC*=94.32 \\ AUC*=75.93 \\ AUC*=82.43} & \makecell{NA} & \multirow{2}{*}{2022} \\
        \hline
        \multirow{2}{*}{Zhao et al.~\cite{SST}} & \makecell{FaceForensics++(HQ)~\cite{Mesoscopic90},\\FaceForensics++(LQ)~\cite{Mesoscopic90}, \\ Celeb-DF~\cite{li2020celebdf}, \\ DFDC~\cite{dolhansky2020deepfake}} & \makecell{ACC=95.33 \\ ACC=83.53\\ACC=99.18 \\ ACC=97.67 } & \makecell{NA}
 & \multirow{2}{*}{2022} \\
        \hline     
        \multirow{2}{*}{GenConViT~\cite{GENCONVIT}} & \makecell{DFDC~\cite{dolhansky2020deepfake}, \\ FaceForensics++~\cite{Mesoscopic90}, \\ TIMIT~\cite{TIMIT}, \\  Celeb-DF-v2~\cite{li2020celebdf}} & \makecell{ACC=98.50 ,AUC=99.90\\ ACC=97.00,AUC=99.6 \\ ACC=98.28 \\ ACC=90.94} & \makecell{https://github.com/erprogs\\/genconvit \\} & \multirow{2}{*}{2023} \\
        \hline     
        \multirow{1}{*}{David et al.~\cite{Coccomini_2022}} & \makecell{DFDC~\cite{dolhansky2020deepfake} } & \makecell{AUC=95.10} & \makecell{https://github.com/davide-coccomini\\} & \multirow{1}{*}{2021} \\
        \hline
    \end{tabular}
    \end{adjustbox}
    \caption{Summary of ViT based deepfake detection methods.* means cross-datasets evaluation}
    \label{tab:sample}
\end{table*}

\section{Vision Transformer in Deepfake Detection}

When it comes to the task of deepfake detection using ViTs, various methods can be categorized into two broad categories: ViT as standalone models and ViT combined with other techniques as hybrid models. Hybrid models can be further classified into sequential models and parallel models based on how the ViT is combined. The classification of the main models is illustrated in Fig.~\ref{overview}, with additional details about each model are summarized in Table.~\ref{tab:sample}.

\begin{figure}[tb]
    \centering
    \includegraphics[width=0.8\linewidth]{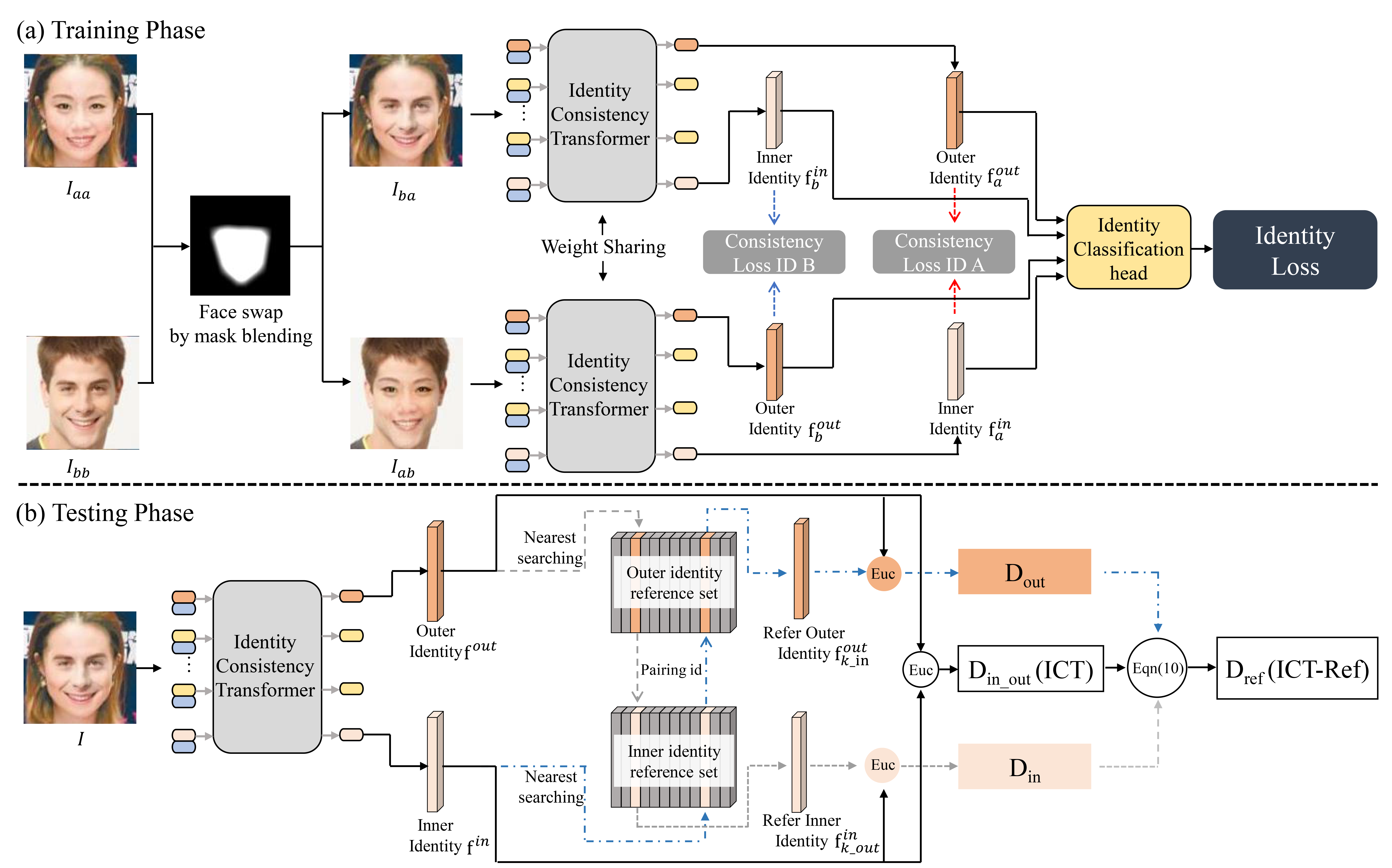}
    \caption{\textbf{Overview architecture of ICT.} (a) Training phase and (b) Testing phase. (The figure is taken from~\cite{ICT})}
    \label{fig:ICT_train}
\end{figure}

\subsection{Standalone Models}
In this approach, a ViT model is trained specifically for deepfake detection. The ViT is trained on a large dataset of real and deepfake images to learn the distinguishing patterns and features that can differentiate between authentic and manipulated content. This method leverages the self-attention mechanism of ViTs to capture spatial relationships within the image, enabling the detection of anomalies or inconsistencies introduced by deepfake techniques.

Dong et al. in~\cite{ICT} proposed Identity Consistency Transformer (ICT), a novel face forgery detection method that focuses on high-level semantics, specifically identity information, and detecting a suspect face by finding identity inconsistency in inner and outer face regions. Fig.~\ref{fig:ICT_train} presents the training and testing phases of ICT. While testing, identity reference sets are used to enhance the identity detection of certain celebrities. Trained on the MS-Celeb-1M~\cite{guo2016msceleb1m} dataset, ICT achieved 98.56\%, 93.17\%, 96.41\%, 94.43\%, and 99.25\% AUC on FF++~\cite{Mesoscopic90}, DFD~\cite{DFD}, Celeb-DF-v1~\cite{li2020celebdf}, Celeb-DF-v2~\cite{li2020celebdf}, and Deeper~\cite{jiang2020deeperforensics10} datasets, respectively. It achieves state-of-the-art performance not only across different datasets but also across various types of image degradation forms found in real-world applications including deepfake videos. ICT presents several merits: (i) It does not rely on any particular facial forgery method, rather assumes the existence of identity inconsistency, showcasing robust generalization capabilities. (ii) It can utilize publicly accessible genuine facial images as a reference set, thereby improving detection performance, particularly in scenarios involving facial forgeries of celebrities. (iii) It exhibits resilience against diverse forms of image degradation, including scaling, noise, and video encoding, making it well-suited for real-world applications. Despite its many advantages, the method has one major drawback: the method mainly detects fake faces with inconsistent identities, and may not be able to detect facial reproduction results with consistent identities.


ICT demonstrates that intra-frame inconsistency is very effective for deepfake detection, but requires additional pixel-level forged location annotations. To acquire such annotations, some existing methods generate large-scale synthesized data with location annotations, which are only composed of real images and cannot capture the properties of forgery regions. Others generate forgery location labels by subtracting paired real and fake images, yet such paired data is difficult to collect and the generated label is usually discontinuous.
\begin{figure}
    \centering
    \includegraphics[width=1\linewidth]{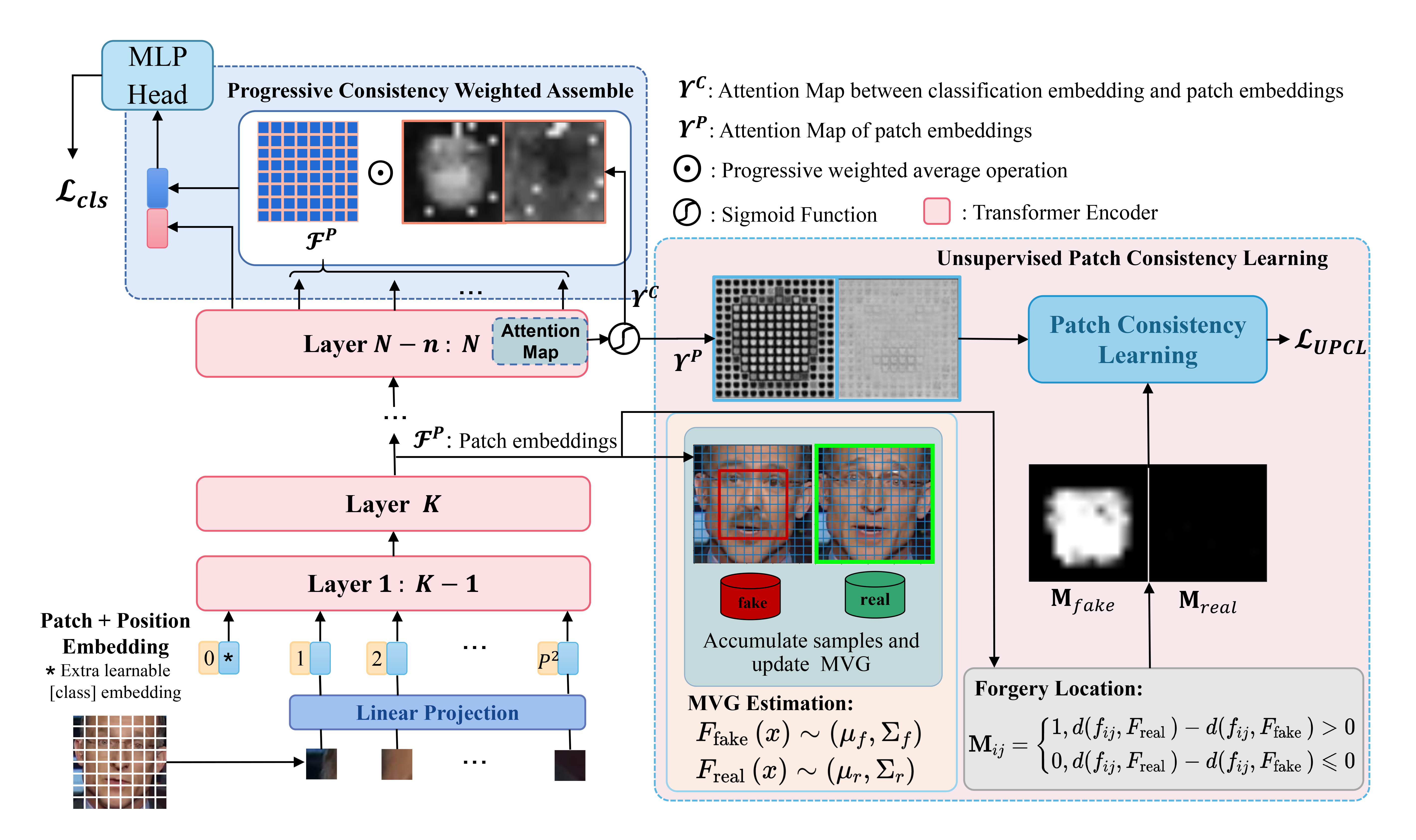}
    \caption{\textbf{Overview architecture of UIA-ViT.} (The figure is taken from~\cite{UIA-ViT})}
    \label{fig:UIA}
\end{figure}
To address these issues, Zhuang et al. proposed a novel Unsupervised Inconsistency-Aware method based on Vision Transformer (UIA-ViT) in~\cite{UIA-ViT}. As shown in Fig.~\ref{fig:UIA}, UIA-ViT has two key components: UPCL (Unsupervised Patch Consistency Learning) and PCWA (Progressive Consistency Weighted Assemble). The process consists of four main steps: (i) Feature Extraction: Employing ViT for facial image feature extraction, the method decomposes the image into a series of blocks and utilizes self-attention mechanisms to learn relationships among these blocks. (ii) Unsupervised Forgery Localization Estimation: Utilizing Multivariate Gaussian (MVG) estimation to represent features of genuine and forged image blocks, and generating pseudo-labels. By comparing the Mahalanobis distance between the features of image blocks and genuine or forged distributions, this approach approximates the location of the forged region without requiring pixel-level annotations. (iii) Block Consistency Learning: Leveraging the self-attention mechanism in ViT to acquire consistency relationships among features of different blocks. Through the design of a consistency loss function, this method constrains attention maps across different layers to capture internal inconsistencies in forged images. (iv) Progressive Consistency Weighted Combination: Generating consistency-aware features using ViT's classification embeddings and block embeddings. By introducing a variable weighting function, this approach progressively combines classification embeddings and block embeddings, thereby gradually enhancing the final detection performance. Similar to ICT, UIA-ViT evaluates detection performance across datasets. Trained on FF++, UIA-ViT gained 99.33\%, 94.68\%, 82.41\%, 86.59\%, 75.80\% AUC on FF++~\cite{Mesoscopic90}, DFD~\cite{DFD}, Celeb-DF-v2~\cite{li2020celebdf}, Celeb-DF-v1~\cite{li2020celebdf}, DFDC~\cite{dolhansky2020deepfake}, respectively. The strengths of UIA-ViT lie in its utilization of the self-attention mechanism of the Vision Transformer. This enables effective capture of both local and global information within images. Additionally, it can learn inconsistencies in facial images without needing pixel-level annotations.

In~\cite{Distillation}, an approach combining ViT and distillation learning is presented. Compared to a conventional ViT model, it incorporates not only class token into the input feature vectors for training the network to distinguish between real and forged videos but also introduces distillation tokens for learning knowledge from a teacher network (EfficientNet~\cite{EfficientNet}), which aims to enhance the network's generalization capabilities. The model attained 97.8\% AUC and 91.9 f1 score on DFDC~\cite{dolhansky2020deepfake}, which outperformed the SOTA model.
In contrast to adding modules to ViT, Shaheen Usmani et al.~\cite{Shallow} proposed a shallow ViT for deepfake detection. The model has 16.48 times fewer parameters
and approximately 2.97 times fewer FLOPS than the baseline ViT. The model performs well even with insufficient training data and computational resources and provides better results than existing deep forgery detection methods (92.15\% ACC and 0.92 AUC on RFF dataset, 88.52\% ACC and 0.88 AUC on RFFD dataset). Due to its lightweight and efficiency, the shallow ViT is particularly suitable for deep forgery detection in real-time scenarios such as social media.

\subsection{Hybrid Models (Sequential Structure)}
Hybrid models combine ViTs with other architectures, such as Convolutional Neural Networks (CNNs), to benefit from the strengths of both approaches. The hybrid models may use the initial layers of a CNN for low-level feature extraction, followed by a ViT for higher-level feature learning and detection. This combination allows the model to capture both local details and global context effectively, enhancing the overall detection performance.
\begin{figure}
    \centering
    \includegraphics[width=0.8\linewidth]{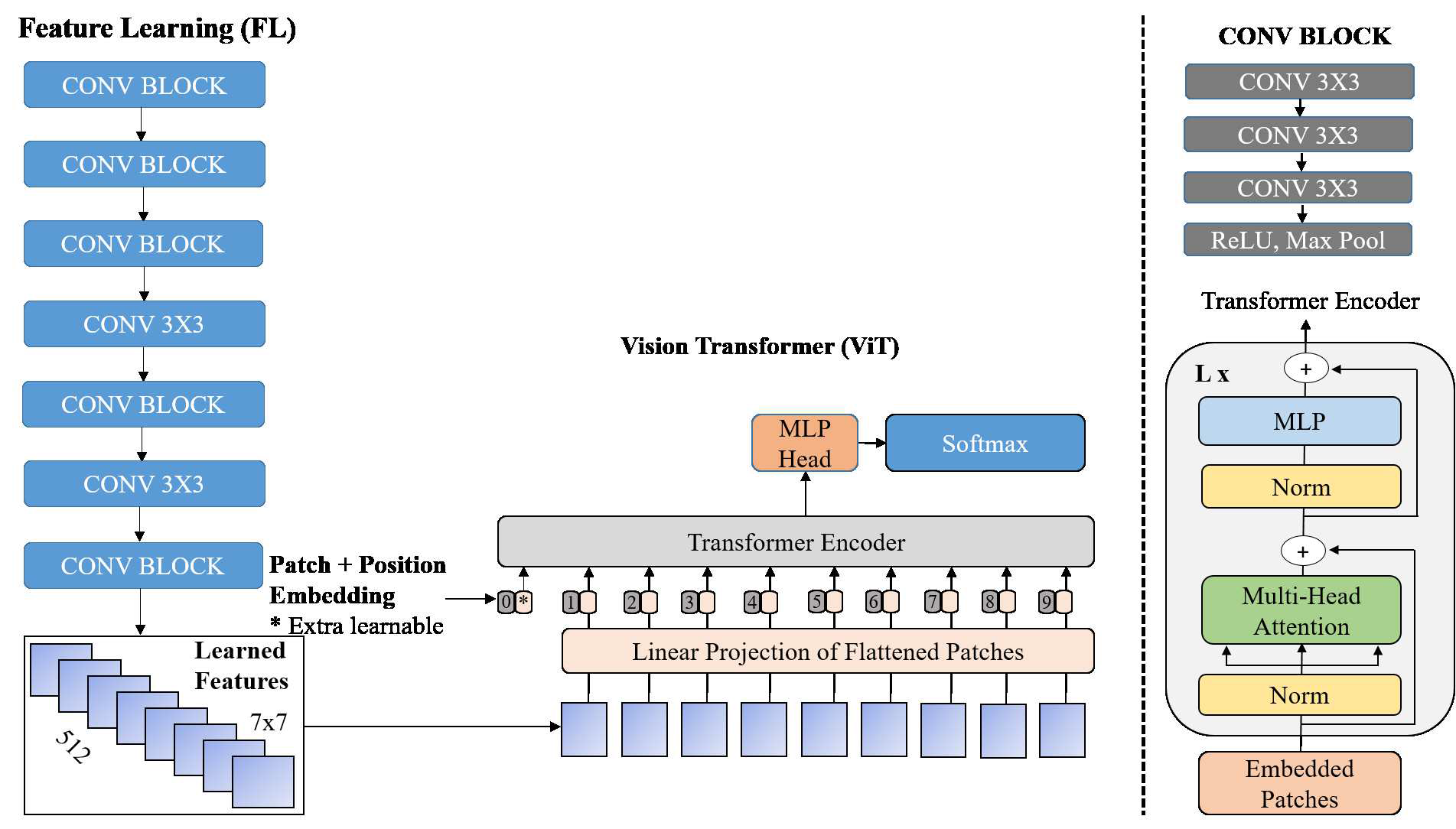}
    \caption{\textbf{Overview architecture of CVIT}  (The figure is taken from~\cite{CVIT})}
    \label{cvit}
\end{figure}

In~\cite{CVIT}, Deressa Wodajo et al. made an initial attempt to combine ViT with CNNs. The proposed Convolutional Vision Transformer (CVIT) is presented in Fig.~\ref{cvit}. It combines the learning capabilities of Convolutional Neural Networks (CNN) and ViTs. CNN excels in learning local features of images, while ViTs can learn both local and global features. This combined capability enables the model to associate with each pixel of the image and comprehend the relationships among non-local features. The CVIT model was trained on a diverse collection of facial images that were extracted from the DFDC dataset and performed well on the DFDC, UADFV, and FaceForensics++ datasets. Nevertheless, the model exhibits suboptimal performance on FF++ FaceShifter dataset. This can be attributed to the inherent difficulty in learning visual artifacts, and it is likely that the proposed CVIT did not effectively capture and understand these artifacts.
\begin{figure}
    \centering
    \includegraphics[width=0.8\linewidth]{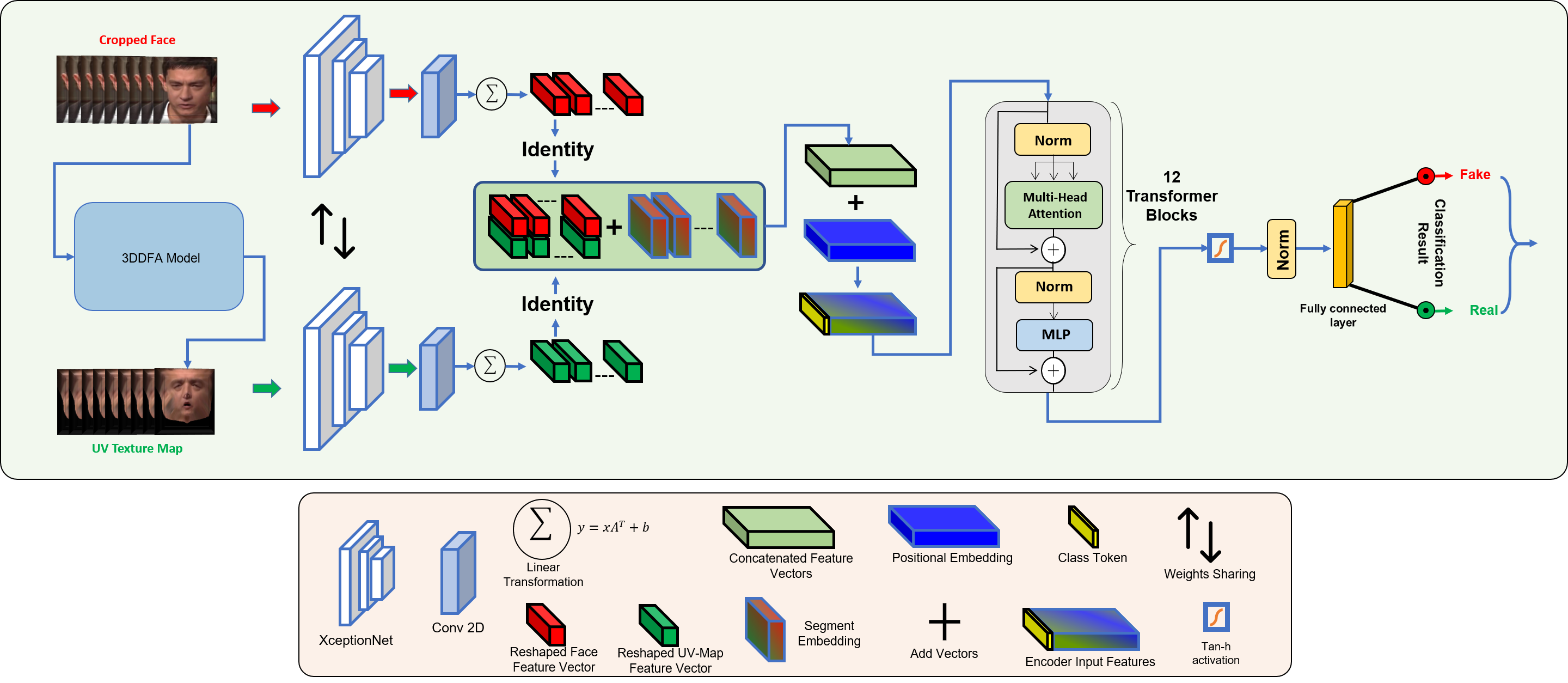}
    \caption{\textbf{Overview architecture of Khan's model.} (The figure is taken from~\cite{Increment})}
    \label{fig:increment}
\end{figure}
Khan et al.~\cite{Increment} proposed a novel video transformer with incremental learning for detecting deepfake videos. As shown in Fig.~\ref{fig:increment}, the proposed model utilizes both the original facial image and the UV texture map for feature extraction. The UV texture map serves the purpose of preventing information loss when the facial orientation is not frontal. Aligning facial images to the UV map provides information about poses, blinking, and mouth movements. Extracting features from both the facial image and its corresponding UV texture map offers a more comprehensive and precise approach to capturing and utilizing facial features. In addition, an incremental learning strategy was adopted to fine-tune the proposed model on new datasets without sacrificing its performance on previous datasets. Following incremental learning on seven distinct datasets, the model exhibited outstanding performance with accuracy scores of 99.79\%, 99.28\%, and 91.69\% on the FF++~\cite{Mesoscopic90}, DFD~\cite{DFD}, and DFDC~\cite{dolhansky2020deepfake} datasets, respectively. It is worth noting that they conducted corresponding ablation experiments. The experimental results indicate that under identical conditions, the hybrid model outperformed the standalone ViT where the former achieved 99.28\% AUC, 98.92\% F1-Score, and 99.28\% ACC on FF++, while the latter only attained 77.10\% AUC, 68.71\% F1-Score, and 73.26\% ACC.

Wang et al.~\cite{Pooling} not only combined ViT with CNN, but also made some improvements to ViT on this basis. A CNN with a kernel size identical to the feature map dimensions considers convolutional computations involving all feature blocks and their relationships. However, such operations reduce the feature map to one dimension, significantly discarding crucial features and diminishing model performance. The utilization of a pooling transformer is employed to adjust the feature dimensions for image analysis.The proposed Deep Convolutional Pooling Transformer initially undergoes feature extraction through CNN convolution. The extracted features are then input into depthwise separable convolution to obtain Q, K, V, which are subsequently fed into the pooling transformer. The model was trained in a self-made keyframe dataset based on FF++~\cite{Mesoscopic90}, cross-dataset evaluated on DFDC~\cite{dolhansky2020deepfake}, Celeb-DF~\cite{li2020celebdf}, DF-1.0~\cite{jiang2020deeperforensics10}. Researchers discovered that deep ViT models may encounter the issue of attention collapse, where attention maps become excessively similar, leading to a noticeable performance decline with increasing model depth. To address this, re-attention mechannism is employed to preserve attention map diversity.

In~\cite{ISTVT}, a video-level deepfake detection model was proposed. ISTVT (Interpretable Spatial-Temporal Video Transformer), which consists of a novel decomposed spatial-temporal self-attention and a self-subtract mechanism to capture spatial artifacts and temporal inconsistency, demonstrated strong performance on multiple datasets. MARLIN~\cite{cai2023marlin} itself is not related to ViT, but their proposed feature extractor MARLIN combined with ViT achieved SOTA performance in FF++~\cite{Mesoscopic90}, which is much better than MARLIN combined with CNNs.

\subsection{Hybrid Models (Parallel Structure)}
Some existing methods employ attention mechanisms to fuse information from both ViTs and other models. This approach combines the strengths of ViTs in capturing global relationships with other models that excel at detecting local artifacts or inconsistencies. Attention-based fusion methods dynamically weigh the contributions of different models or features based on their relevance, enhancing the overall deepfake detection performance.

Davide et al.~\cite{Coccomini_2022} presented a deepfake detection model combining EfficientNet and ViTs. In contrast to the current state-of-the-art methodologies, they do not employ distillation or ensemble techniques. The top-performing model attained an AUC of 0.951 and an F1 score of 88.0\%, which is in close proximity to the current state-of-the-art performance on the DFDC~\cite{dolhansky2020deepfake} dataset.

\begin{figure}
    \centering
    \includegraphics[width=0.8\linewidth]{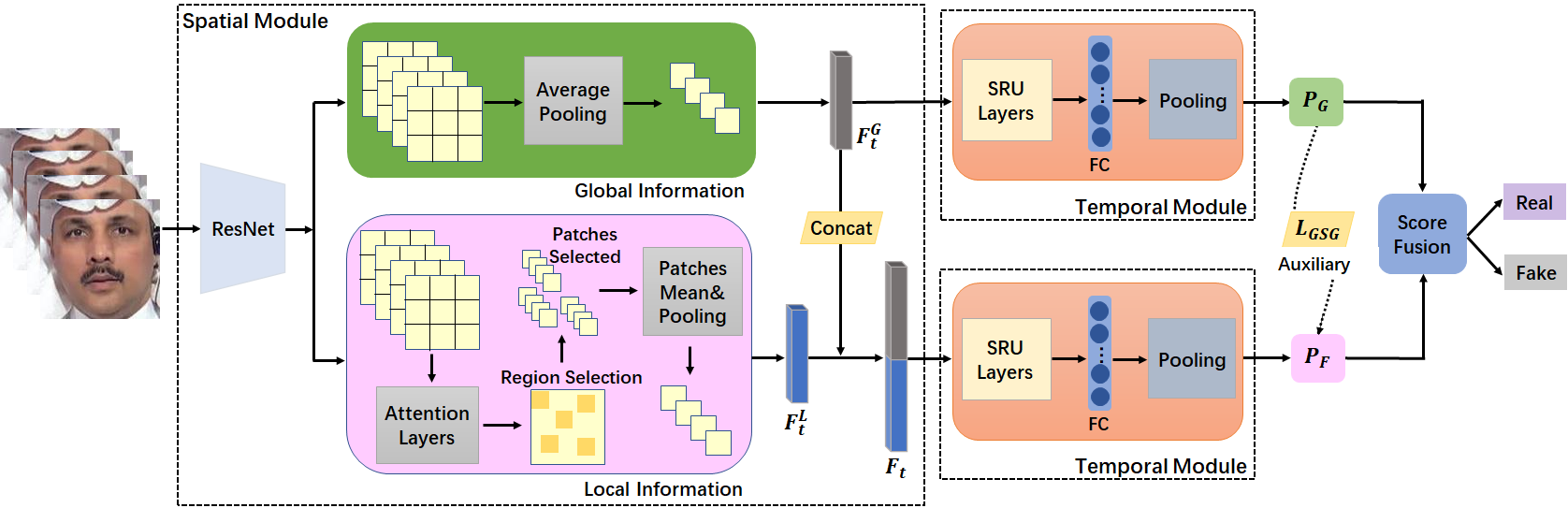}
    \caption{\textbf{Overview architecture of Zhao' model.} (The figure is taken from~\cite{Spatiotemporal94})}
    \label{zhao}
\end{figure}

In~\cite{Spatiotemporal94}, Zhao et al. proposed a novel model consist of a spatial module and a temporal module as shown in Fig.~\ref{zhao}. The spatial module comprises global information flow and local information flow. The input frame sequences undergo feature extraction through these two streams, and the extracted features, after fusion, are then input into the temporal module to capture temporal detection information. The attention layer of the ViT is applied to the local part of the spatial module. They also designed a novel regularization loss called the Global Stream Guidance (GSG) loss, used to guide the selection of local information and the extraction of temporal information in the fusion stream. This loss facilitates the comprehensive utilization of the inherent complementary advantages of both global and local information. Intra-dataset and cross-dataset evaluations were conducted on FF++~\cite{Mesoscopic90}, DFDC~\cite{dolhansky2020deepfake}, and Celeb-DF~\cite{li2020celebdf}, resulting in SOTA performance. Furthermore, they conducted ablation studies on the sampling frame numbers and model complexity, arriving at a conclusion similar to~\cite{Pooling}: for deep ViT models, deeper and larger does not necessarily lead to better performance.

\begin{figure}
    \centering
    \includegraphics[width=0.8\linewidth]{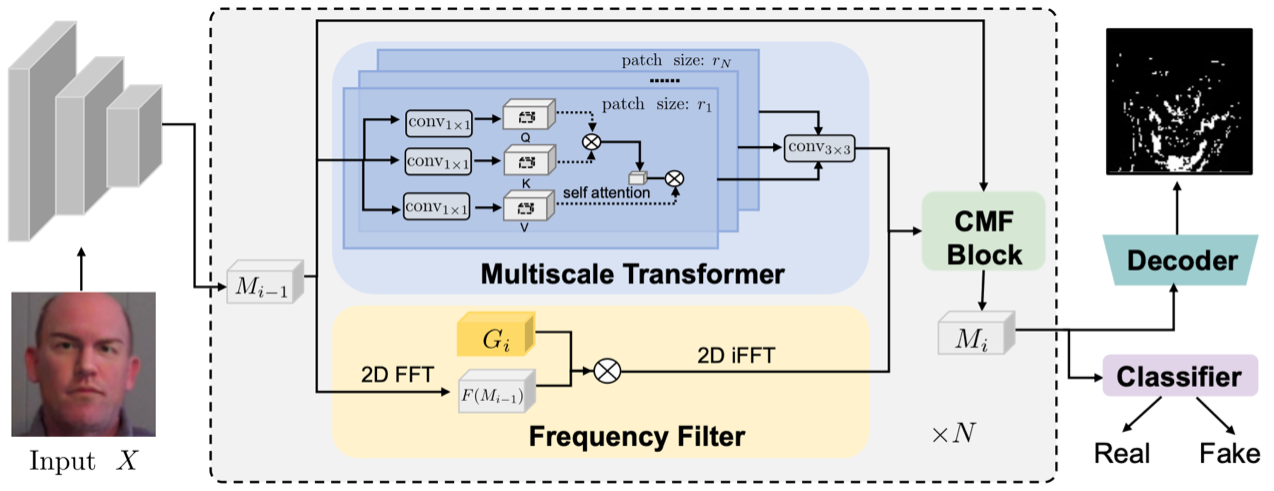}
    \caption{\textbf{Overview architecture of M2TR.} (The figure is taken from~\cite{M2TR})}
    \label{fig:M2TR}
\end{figure}

Xue et al.~\cite{FOFDTD} introduced a DeepFake detection method specifically tailored for subtle facial expression manipulation, facial detail alterations, and blurred images. The proposed method framework includes an organ-selection module, a facial-region interception module, an organ-level transformer, and a classifier. The organ-level transformer identifies deepfake features through organs, utilizing an organ-selection module to reduce the weight of compromised, damaged, and low-quality organs, thereby enhancing accuracy. Simultaneously, a full-face ViT is employed to assist in detecting partial information. The accuracy experiences significant improvement when an increased number of key organ features are used. This method has achieved outstanding performance on numerous datasets, particularly reaching SOTA results on datasets with low-resolution data for its organ detection doesn't rely on high-resolution data.

Generative Convolutional Vision Transformer (GenConViT) was proposed in~\cite{GENCONVIT}. GenConViT comprises two independently trained networks and four main modules: Autoencoder (AE), Variational Autoencoder (VAE), ConvNeXt layer, and Swin Transformer. The first network employs AE to generate the latent feature (LF) space of input images, maximizing the probability of category predictions, i.e., determining whether the given input is a deepfake. The second network uses VAE to reconstruct images while simultaneously maximizing category prediction probability and minimizing the loss distance between sample images and reconstructed images. The AE and VAE models extract LF from input video frames, capturing hidden patterns and correlations in the learned visual artifacts of deepfake manipulation. GenConViT achieved classification accuracies of 98.5\%, 98.28\%, 97\%, and 90.94\% on DFDC~\cite{dolhansky2020deepfake}, TIMIT~\cite{TIMIT}, FF++~\cite{Mesoscopic90}, and Celeb-DF~\cite{li2020celebdf}, respectively, demonstrating its robust performance.

In~\cite{M2TR}, the authors proposed a Multi-modal Multi-scale Transformer (M2TR), which operates on patches of different sizes to detect local inconsistencies in images at different spatial levels. As presented in Fig.~\ref{fig:M2TR}, the approach aims to capture the subtle manipulation artifacts at different scales by using transformer models. Apart from RGB information, M2TR further learns to detect forgery artifacts in the frequency domain through a carefully designed cross modality fusion block. The combination of multiscale transformer and frequency filter enables M2TR to extract effective features from highly compressed images. This capability results in outstanding performance across three different resolution datasets in FF++, particularly excelling in low-resolution datasets compared to other methods. Ablation experiments in the paper indicate that the multiscale transformer is the most crucial component of M2TR. The absence of the multiscale transformer module leads to a significant performance drop across various datasets, particularly in low-quality datasets, with a decrease in accuracy approaching 6\%.

\section{Benchmark}
As shown in Table~\ref{tab:sample}, these models have demonstrated exceptional proficiency on their designated datasets. Nonetheless, given the significant variation in the feature distribution across these datasets, it would be imprudent to directly compare their accuracy metrics to ascertain their relative superiority or inferiority. Therefore, we have decided to reproduce and test models with open-source code on the same datasets. The datasets we decide to use are FaceForensics++(FF++)\cite{Mesoscopic90} and Celeb-DF\cite{li2020celebdf}. 

FaceForensics++ \cite{Mesoscopic90} is a comprehensive benchmark dataset designed for the evaluation of digital face manipulation detection algorithms. It extends its predecessor, FaceForensics, by incorporating a richer and more diverse set of manipulated facial videos. The dataset encompasses a wide array of manipulations, including but not limited to deepfake generation, face swapping, facial reenactment, and more traditional computer graphics-based modifications. We mainly selected 4 different fake videos (Deepfakes, Face2Face, FaceSwap, NeuralTextures), and the original video for training and testing. In addition, FF++ has three different compression qualities, we chose raw to test the performance of the model, and low quality to test the robustness of the model to compressed video.

Celeb-DF\cite{li2020celebdf} is a large-scale, challenging dataset specifically designed for the detection of deepfake videos, with a focus on featuring celebrities. This dataset stands out due to its high-quality deepfake video generation, which significantly reduces common artifacts associated with deepfake content, such as unnatural blinking patterns, facial distortions, and poor lip-syncing. By prioritizing the realism of the deepfakes, Celeb-DF provides a rigorous benchmark for deepfake detection algorithms, aiming to mirror the sophistication and quality of deepfakes encountered in real-world scenarios.

It is worth mentioning that although each model uses the same dataset, in the data preprocessing section, we process the input data of each model according to the respective original paper in order to maximise the reproduction of the best performance of the model. For all these model, we didn't use fine tuning, so the performance could not be great as it in their original implement. But in this way, we can avoid cheery picking and gain a relatively more equitable outcome.

The results are listed in table ~\ref{table:bench}. The CVIT model seems to struggle relative to the others, which could indicate potential areas for improvement, such as feature extraction or model architecture adjustments. Actually GENCVIT is an improved model based on CViT, it achieve much better performance, particularly on the raw FF++ data, which suggests that it may be better at handling less-processed data. M2TR model consistently shows high performance across all three datasets in both accuracy (ACC) and area under the receiver operating characteristic curve (AUC) metrics, indicating robustness and generalizability. Most of the models show large performance degradation when faced with compressed video. Most of the models show large performance degradation when faced with compressed video. However, khan et al~\cite{Increment} and M2TR~\cite{M2TR} still maintain good performance, the former by UV texture map and the latter by frequency domain feature extraction.

\begin{table}[ht]
\centering
\begin{tabular}{|l|l|l|l|}
\hline
Model & FF++(LOW) & FF++(RAW) & Celeb-DF \\ \hline
David et al~\cite{Coccomini_2022} & \begin{tabular}[c]{@{}l@{}}50.71\% ACC\\ 0.763 AUC\end{tabular} & \begin{tabular}[c]{@{}l@{}}81.85\% ACC\\ 0.902 AUC\end{tabular} & \begin{tabular}[c]{@{}l@{}}81.27\% ACC\\ 0.870 AUC\end{tabular} \\ \hline
Khan et al ~\cite{Increment}& \begin{tabular}[c]{@{}l@{}}77.14\% ACC\\ 0.723 AUC\end{tabular} & \begin{tabular}[c]{@{}l@{}}86.44\% ACC\\ 0.907 AUC\end{tabular} & \begin{tabular}[c]{@{}l@{}}73.78\% ACC\\  0.843 AUC\end{tabular} \\ \hline
CVIT~\cite{CVIT} & \begin{tabular}[c]{@{}l@{}}58.78\% ACC\\ 0.647 AUC\end{tabular} & \begin{tabular}[c]{@{}l@{}}69.16\% ACC\\ 0.675 AUC\end{tabular} & \begin{tabular}[c]{@{}l@{}}70.27\% ACC\\ 0.679 AUC\end{tabular} \\ \hline
GENCVIT~\cite{GENCONVIT} & \begin{tabular}[c]{@{}l@{}}48.56\% ACC\\ 0.884 AUC\end{tabular} & \begin{tabular}[c]{@{}l@{}}97.68\% ACC\\ 0.997 AUC\end{tabular} & \begin{tabular}[c]{@{}l@{}}90.95\% ACC\\ 0.981 AUC\end{tabular} \\ \hline
M2TR~\cite{M2TR} & \begin{tabular}[c]{@{}l@{}}87.19\% ACC\\ 0.904 AUC\end{tabular} & \begin{tabular}[c]{@{}l@{}}95.82\% ACC\\ 0.987 AUC\end{tabular} & \begin{tabular}[c]{@{}l@{}}98.46\% ACC\\  0.999 AUC\end{tabular} \\ \hline
\end{tabular}
\caption{Benchmark over 3 datasets}
\label{table:bench}
\end{table}

\section{Open Issues \& Future Work}
In this section, we investigate open challenges and future research directions regarding the application of vision transformers in deepfake detection.
\subsection{Open Issues}

\subsubsection{Model Drift}
Existing deepfake detection systems face challenges in keeping pace with the continuous evolution of deepfake techniques. The frozen nature of these systems upon deployment results in a growing gap between their capabilities and the rapidly advancing methods employed by deepfake generators. There is a pressing need to develop adaptive detection mechanisms that can dynamically evolve to counter emerging deepfake strategies, ensuring ongoing effectiveness in the ever-changing landscape of deepfake.
\subsubsection{Data Scarcity and Quality}
ViT-based deepfake models require extensive training data, particularly high-quality data of the target individual's face or voice. The limited availability of high-quality data can hinder the training process. For instance, generating deepfakes of lesser-known individuals or private individuals may be challenging due to the scarcity of suitable training data. Additionally, the quality and diversity of training data significantly impact the realism and diversity of generated content. A lack of diverse training data can lead to biased or unrealistic deepfakes.
\subsubsection{Temporal Consistency}
Maintaining temporal consistency, especially in video deepfakes, remains a challenge. Transformers generate content frame-by-frame, and ensuring that each frame seamlessly transitions from the previous one is complex. Any temporal inconsistency, such as abrupt facial movements or unnatural lip synchronization, can be a telltale sign of a deepfake. Techniques for improving temporal coherence, such as recurrent neural networks (RNNs) or Long Short-Term Memory (LSTM) structures, are still being researched and integrated with transformers to address this issue.
\subsubsection{Bias and Fairness}
Transformer models, including those used in deepfakes, can inherit biases from their training data. This bias can manifest in the generated content and reinforce societal prejudices. Techniques for bias mitigation, such as fairness-aware training and data preprocessing, are being researched. Achieving fairness and mitigating bias in transformer-based deepfakes requires a nuanced understanding of the underlying biases and careful model design to promote equitable content generation.

\subsection{Future work}
\subsubsection{Improving Model Explainability}
Typically, in the existing literature, the majority of deep-learning-based deepfake detection techniques fail to provide a comprehensive explanation for the final outcome of their detection process. Enhancing interpretability is crucial for researchers to understand how ViT models make decisions in deepfake detection and adjust the model according to the outcome.

\subsubsection{Improving Model Generalization}
ViT-based deepfake detection models may struggle to generalize well across diverse deepfake datasets. It can be seen from Table 1 that most of the existing methods' performances decrease remarkably when evaluated cross datasets. New techniques such as incremental learning and transfer learning will be researched to enhance generalization and robustness of models.

\subsubsection{Enhancing Multi-Modal Approaches}
Detecting deepfakes is mainly based on visual information. A few models also use temporal and frequency domain information. More robust deepfake detection should consider multiple modalities, investigating the fusion of audio, text, or contextual information with visual cues. There is significant research scope for developing novel architectures for multimodal deep fake detection.

\subsubsection{Benchmarking and Evaluation Standards}
In terms of deepfake research, many of the methods are not published in open source code and hard to reproduced. Direct comparison of the performance of different models is inaccurate due to differences in experimental environments and model deployment. The promotion of reproducible results should be encouraged through the provision of extensive datasets, experimental configurations, and open-source codes. Establish comprehensive benchmarks and standards for evaluating deepfake detection is essential for future development.

\section{Conclusion}
This survey has provided an in-depth exploration of the technical intricacies and advancements within the realm of transformer-based deepfake detection technology. Our exploration of this intricate landscape has unveiled the remarkable capabilities of ViT-based models in identifying forged content across diverse modalities. It is clear that transformers have brought about a paradigm shift in the field, showcasing cutting-edge performance and unparalleled versatility. The survey also delves into open issues and future directions, providing a roadmap for further research. Additionally, we present detailed information on the datasets employed, performance metrics, and access links to the source codes of each discussed model. It is hoped that this survey will inspire researchers who are interested in further integrating ViT into deepfake detection, providing them with valuable insights and convenience in their endeavors.





\bibliographystyle{elsarticle-num}

\bibliography{main.bib}

\end{document}